\documentclass[10pt,twocolumn,letterpaper]{article}

\usepackage[pagenumbers]{wacv} 

\definecolor{wacvblue}{rgb}{0.21,0.49,0.74}
\usepackage[pagebackref,breaklinks,colorlinks,allcolors=wacvblue]{hyperref}

\usepackage{CJKutf8}
\usepackage{longtable}
\usepackage{multirow}
\usepackage{booktabs}

\usepackage{comment}

\def\wacvPaperID{***} 
\def\confName{WACV}
\def\confYear{2027}

\title{JSL-DC: A Word-Level Japanese Sign Language Dataset with Linguist-Derived Descriptions for Distinguishing Confusable Signs}

\author{
Ken Takaki$^{1}$\quad Asuka Ando$^{1}$\quad Misa Suzuki$^{2}$\quad Uiko Yano$^{3}$\quad
Masaya Tsujimoto$^{1}$\quad Bill Neubauer$^{4}$\\
Ananay Vikram Gupta$^{4}$\quad Rose Shao$^{5}$\quad Matthias Hoppe$^{5}$\quad
Sahir Shahryar$^{4}$\quad Celeste Mason$^{4}$\\
Kai Kunze$^{6}$\quad Yohei Oseki$^{1}$\quad Yoshihiro Kawahara$^{1}$\quad Thad Starner$^{4}$\\[0.6em]
$^{1}$The University of Tokyo\quad
$^{2}$Independent\quad
$^{3}$Kwansei Gakuin University\\
$^{4}$Georgia Institute of Technology\quad
$^{5}$Keio University\quad
$^{6}$Technical University of Clausthal\\[0.4em]
{\tt\small takaki@akg.t.u-tokyo.ac.jp}
}
\begin{document}
\maketitle
\begin{abstract} 
Effective sign language (SL) acquisition is crucial for deaf children, yet 95\% are born to hearing parents who often lack proficiency in SL. SL recognition can power learning tools to help parents communicate with their children. However, Japanese Sign Language (JSL) lacks large-scale, multi-signer datasets, hindering the development of models that can generalize to new users.
To address this gap, we introduce JSL-DC, the largest JSL dataset by video count, comprising 36.7K videos from 19 signers. 
The entire process was Deaf-centric: the lexicon comprising 270 JSL words was selected by Deaf and Coda linguists to facilitate parent-child communication, all participants were Deaf individuals who use JSL daily, and the data underwent a two-stage review process involving Deaf linguists. 
Moreover, we provide linguist-derived descriptions for distinguishing confusable signs.
We demonstrate that the proposed model inspired by the descriptions outperforms state-of-the-art recognition methods by 9.8\% on the confusable subset. The dataset, along with its linguistic description that inspires new models, will be released under a CC-BY 4.0 license to accelerate research in SL recognition. 
\end{abstract}

\section{Introduction}
\label{sec:intro}

Sign language is a native language for many deaf or hard-of-hearing individuals, and its acquisition during early childhood is crucial for language development~\cite{Lederberg1998-yn}. Parent-child communication is vital for this process. However, 95\% of deaf children are born to hearing parents, who often struggle to communicate effectively as the parents do not know sign language~\cite{Moores2001-rm}. Currently, learning sign language poses a significant barrier for these parents~\cite{Neubauer2026-ru}, partly because access to instructions is limited and often requires traveling long distances, sometimes more than an hour. 
Early language acquisition in children often begins with short utterances that combine pointing and single words~\cite{Takei2001-tx}; therefore, allowing parents to even acquire a basic vocabulary, without full fluency, can meaningfully support early communication.
While online videos offer self-study options, learning is considered more effective when combined with hands-on practice~\cite{Starner2023-vt}. 
An isolated sign language recognition (ISLR) based ``bubble bursting game'' has been proposed as a method to facilitate hands-on practice of sign language words~\cite{Starner2023-vt}.

Large-scale ISLR datasets are available for many sign languages (\Cref{tab:dataset-comparison}), including American (PopSign ASL v1.0~\cite{Starner2023-vt}: 200K videos, WLASL~\cite{Li2019-px}: 21K videos), British (BSL-1K~\cite{Albanie2020-wp}: 273K videos), Chinese (CSL500~\cite{Huang2019-kf}: 125K videos, NationalCSL-DP~\cite{Jin2025-nk}: 134K videos), Australian (MM-WLAuslan~\cite{Shen2024-tz}: 283K videos), and Turkish (AUTSL~\cite{Sincan2020-xh}: 38K videos) sign languages. Starting with datasets like WLASL~\cite{Li2019-px}, numerous ISLR models have been proposed. Consequently, recognition accuracy has reached practical levels for ASL~\cite{Zuo2023-qs, Hruz2022-dp, Hu2023-oz}, Auslan~\cite{Shen2024-tz}, CSL~\cite{Huang2019-kf}, and TSL~\cite{Jiang2021-og}, often exceeding 80\% accuracy on classification tasks with over 100 glosses.
A critical limitation remains, however. Some sign pairs share a handshape and differ only in non-manual cues, such as mouth or facial movements, or in how they are used in context. Datasets rarely label which pairs are confusable, and to our knowledge, none explain how to tell each pair apart, leaving computer vision researchers without linguistic guidance on these hard cases.


Moreover, datasets for Japanese Sign Language (JSL) are extremely scarce. One available dataset contains over 6,000 signs but is limited to a single signer~\cite{Nagashima2021-ni}, making it almost impossible to build ISLR models that generalize to new users. Another dataset includes 121 signers but only 100 signs, offering a very limited vocabulary~\cite{Bono2014-dg}. As a result, the development of JSL recognition models has lagged significantly. Due to the reliance on single-signer data~\cite{Nagashima2021-ni, Osugi2012-zw}, some studies fail to evaluate on unseen signers~\cite{Ando2024-ow, Bai2024-yw}. Others that do perform unseen-signer evaluation use data from novice signers imitating sign motions~\cite{Takazume2023-os}, and their performance is expected to degrade when applied to native signers.

In this paper, we introduce JSL-\textbf{DC}, the largest JSL dataset by video count, exceeding the previous largest by 3 times and comprising 36.7K videos from 19 signers. It covers 270 signs with 136 repetitions per sign.
We outline our contribution as follows. 
\begin{itemize}
    \item \textit{\textbf{D}eaf \textbf{C}entric data collection}: the lexicon for facilitating the communication between hearing parents and deaf children was determined by Deaf and child of deaf adults (Coda\footnote{We use Coda instead of CODA, following this author's self-identification.}) linguists. Moreover, all participants in the data collection are \textbf{Deaf individuals who use JSL daily}, and the dataset's quality is ensured by a \textbf{two-stage review process} conducted by reviewers raised by deaf parents and Deaf linguists.
    \item \textit{\textbf{D}istinguishing \textbf{C}onfusable Signs by descriptions by Deaf linguists}: 
    To enhance research on distinguishing glosses that are confusable, JSL linguists annotate pairs of glosses that cannot be distinguished by handshape alone, and we add an explanation of how to distinguish each pair.
    \item \textit{\textbf{D}emonstrating less \textbf{C}onfusion and 9.8\% F1 score gain}: 
    Based on the descriptions added by the Deaf linguist, we resolve the confusion among glosses that are distinguished by mouth movement using AV-HuBERT used in lip-reading tasks.
    As a result, we show a further 9.8\% F1 score gain on the confusable glosses over a state-of-the-art model that already uses facial features, demonstrating that linguists' descriptions of how to distinguish signs help guide improvements in sign language recognition.
\end{itemize}

We note that this dataset is not intended for training JSL-to-Japanese translation models, but rather for training recognition models for sign language learning tools; thus, we focused on collecting a specific variant of a sign. 
The \textbf{dataset, including facial data}, will be \textbf{released under a Creative Commons CC-BY 4.0 license} with full participant consent, which we believe will accelerate research in ISLR.


\section{Related Work}
\label{sec:related}

Sign language recognition (SLR) is typically divided into two domains: continuous (CSLR)~\cite{Martinez-Hinarejos2017-si, Wang2025-os, Hu2023-ah, Papadimitriou2023-wi, Zuo2022-pl, Hu2023-oz} and isolated (ISLR) recognition~\cite{Neubauer2026-ru, Zuo2023-qs, Hruz2022-dp, Hu2023-oz, Jiang2021-og, Bohacek2022-bp, Hu2024-zo, Martinez-Hinarejos2017-si, Zhao2024-kv}, which focus on sentence-level and word-level videos, respectively.

ISLR is central for sign language learning applications because it provides the word-level recognition needed for vocabulary practice. To support JSL learning tools, this work introduces a new ISLR dataset for Japanese Sign Language. Accordingly, this section provides an overview of the ISLR landscape, covering datasets, recognition methodologies, and recent trends, with an emphasis on Japanese Sign Language (JSL).

\subsection{Datasets for Isolated Sign Language Recognition}
\begin{table*}[ht!]
\centering
\footnotesize
\caption{Comparison of existing ISLR datasets. Distinguish Note is a description by a Deaf linguist for distinguishing confusable signs.}
\label{tab:dataset-comparison}
\begin{tabular}{@{}lccccccc@{}}
\toprule
\textbf{Dataset} & \textbf{Country} & \textbf{Signs} & \textbf{Signers} & \textbf{Videos} & \textbf{Ave. Videos/Sign} & \textbf{Source} & \textbf{Distinguish Note} \\
\midrule
Purdue RVL-SLLL \cite{Martinez2003-rz} & USA & 39 & 14 & 2.6K & 67 & Studio & \\
RWTH-BOSTON 50 \cite{Zahedi2005-by} & USA & 50 & 3 & 0.5K & 10 & Studio & \\
ASLLVD \cite{Athitsos2008-ni} & USA & 3,000 & 6 & 9.8K & 3 & Studio & \\
WLASL \cite{Li2019-px} & USA & 2,000 & 119 & 21.1K & 11 & Web & \\
MS-ASL \cite{Joze2018-ud} & USA & 1,000 & 222 & 25.5K & 26 & Web & \\
ASL Citizen \cite{Desai2023-po} & USA & 2,731 & 52 & 83.9K & 31 & Webcam & \\
PopSign ASL v1.0 \cite{Starner2023-vt} & USA & 250 & 47 & 200.7K & 803 & Phone & \\
PopSign ASL v2.0 \cite{Neubauer2026-ru} & USA & 562 & 76 & 363.7K & 647 & Phone & \\
BOBSL \cite{Albanie2021-xj} & GBR & 2,281 & 39 & 452K & 198 & TV & \\
 BSL-1K \cite{Albanie2020-wp} & GBR & 1,064 & 40 & 273.0K & 257 & TV & \\
BSLDict \cite{Momeni2020-qw} & GBR & 9283 & 148 & 14.2K & 2 & Studio & \\

DEVISIGN-D \cite{Chai2012-rj} & CHN & 500 & 8 & 6K & 12 & Studio & \\
DEVISIGN-G \cite{Chai2012-rj} & CHN & 36 & 8 & 432 & 12 & Studio & \\
DEVISIGN-L \cite{Chai2012-rj} & CHN & 2,000 & 8 & 24.0K & 12 & Studio & \\
CSL 500 \cite{Huang2019-kf} & CHN & 500 & 50 & 125.0K & 250 & Studio & \\
NationalCSL-DP \cite{Jin2025-nk} & CHN & 6,707 & 10 & 134.1K & 20 & Studio & \\
UWB-06-SLR-A \cite{Campr2007-dz} & CZE & 25 & 15 & 5.5K & 220 & Studio & \\

DGS Kinect 40 \cite{Ong2012-cs} & DEU & 40 & 14 & 2.8K & 70 & Studio & \\
SMILE \cite{Ebling2018-eu} & DEU/CHE & 100 & 30 & -- & -- & Studio & \\

GSL 982 \cite{Ong2012-cs} & GRC & 982 & 1 & 4.9K & 5 & Studio & \\

INCLUDE \cite{Sridhar2020-sq} & ISR & 263 & 7 & 4.3K & 16 & Studio & \\
KL\_MV2DSL \cite{Mopidevi2023-sz} & ISR & 200 & 5 & 5.0K & 25 & Studio & \\

LSA64 \cite{Ronchetti2023-yz} & ARG & 64 & 10 & 3.2K & 50 & Studio & \\

LSE-Sign \cite{Gutierrez-Sigut2016-hf} & ESP & 2,400 & 2 & 2.4K & 1 & Studio & \\

LSFB-ISOL \cite{Fink2021-uo} & FRA/BEL & 395 & 100 & 47.6K & 121 & Studio & \\

Logos \cite{Ovodov2025-mt} & RUS & 2,863 & 381 & 200K & 70 & Phone \& Webcam & \\
Slovo \cite{Kapitanov2023-mr} & RUS & 1,000 & 194 & 20K & 20 & Phone \& Webcam & \\

Bosphorus Sign \cite{Camgoz2016-yi} & TUR & 855 & 10 & 51.3K & 60 & Studio & \\
Bosphorus Sign22K \cite{Ozdemir2020-ku} & TUR & 744 & 6 & 22.5K & 30 & Studio & \\
AUTSL \cite{Sincan2020-xh} & TUR & 226 & 43 & 38.3K & 169 & Studio & \\

Auslan-Daily \cite{Shen2023-ik} & AUS & 13,945 & 67 & 25.1K & 2 & Web & \\
MM-WLAuslan \cite{Shen2024-tz}& AUS & 3,215 & 73 & 282.9K & 88 & Studio & \\

\midrule
Colloquial Corpus of JSL \cite{Bono2014-dg} & JPN & 100 & 121 & 12.1K & 121 & Studio & \\
KoSign \cite{Nagashima2021-ni} & JPN & 6359 & 1 & 6.4K & 1 & Studio & \\
A Guide to Hand Gestures and Words \cite{Osugi2012-zw} & JPN & 469 & 1 & 469 & 1 & Studio & \\
\textbf{Ours (reference videos)} & \textbf{JPN} & \textbf{270} & \textbf{2} & \textbf{540} & \textbf{2} & \textbf{Studio} & \textbf{\checkmark}\\
\textbf{Ours} & \textbf{JPN} & \textbf{270} & \textbf{19} & \textbf{36.7K} & \textbf{136} & \textbf{Tablet} & \textbf{\checkmark}\\

\bottomrule
\end{tabular}
\end{table*}


As shown in \Cref{tab:dataset-comparison}, various datasets for ISLR have been proposed worldwide. 
While large-scale datasets comprising over 100,000 videos exist for American (PopSign ASL v1.0~\cite{Starner2023-vt}, PopSign ASL v2.0~\cite{Neubauer2026-ru}), British (BSL-1K~\cite{Albanie2020-wp}), Chinese (CSL500~\cite{Huang2019-kf}, NationalCSL-DP~\cite{Jin2025-nk}), and Australian (MM-WLAuslan~\cite{Shen2024-tz}) sign languages, only NationalCSL-DP flags which of these are confusing signs that cannot be recognized from hand shape alone.
To our knowledge, no dataset provides explicit linguist-authored distinguishing descriptions.
Furthermore, in many datasets, the signers are novices who imitate signs, and annotation is rarely performed by Deaf linguists.
It is therefore questionable whether existing datasets have correctly collected confusing signs in the first place.



Among existing JSL datasets, KoSign~\cite{Nagashima2021-ni} offers a large vocabulary of 6,359 signs, but includes only a single signer, which limits the model's ability to generalize to unseen signers. The Colloquial Corpus of Japanese Sign Language~\cite{Bono2014-dg} features 121 signers but was primarily intended to map regional variations in JSL expressions, resulting in a limited vocabulary of 100. Furthermore, it is unsuitable for ISLR tasks because its videos often contain multiple repetitions of a sign or include unannotated, continuous conversations between Deaf signers. 

In this work, we build the largest JSL dataset by video count, exceeding the previous largest by 3 times~\cite{Bono2014-dg}. We asked 19 participants to record sign data using tablets with our collection app in their daily life environments. Our dataset is the first JSL dataset collected using mobile devices in such a natural setting. 
Furthermore, our dataset includes pairs of confusing signs that cannot be recognized from hand shape alone, together with linguists' explanations of how to distinguish them.
A distinguishing description is only useful if every clip in a confusable class shows the intended sign. 
Enforcing this is hard: even though all our participants use JSL daily, 9.8\% of clips collected for the confusable glosses were reclassified as variants in review, their meaning differing from the reference or not produced correctly. 
Filtering these keeps each class clean, and their prevalence shows that review by Deaf signers is essential for confusable-sign data.

\subsection{Isolated Sign Language Recognition Methods}


ISLR involves classifying the gloss label of a sign. This field is broadly divided into three main approaches: pose-based, pixel-based, and hybrid methods.

Pose-based ISLR methods first extract skeletal information using tools such as MediaPipe~\cite{Google2025-ku} and then process these features with a recognition model~\cite{Neubauer2026-ru, Bohacek2022-bp, Hu2024-zo, Martinez-Hinarejos2017-si, Zhao2024-kv}. Since these methods extract low-dimensional skeletal features using lightweight models, they are generally fast at inference and robust to variations in skin tone and illumination~\cite{Hruz2022-dp}. However, pose estimators like MediaPipe struggle with challenges, such as low accuracy in depth estimation and difficulties with hand occlusions, both of which can lead to substantial performance degradation~\cite{Holmes2024-kp}.

In contrast, pixel-based ISLR processes raw video frames directly using recognition models such as CNNs~\cite{Carreira2017-ne, Tran2015-og}. While these methods are computationally intensive, they tend to achieve higher accuracy in distinguishing between visually similar signs~\cite{Hruz2022-dp}.
Recently, hybrid approaches that combine pose-based and pixel-based features have been introduced to further improve performance~\cite{Zuo2023-qs, Hruz2022-dp, Hu2023-oz, Jiang2021-og, Hu2024-zo}.

ISLR for JSL has often been constrained by datasets limited to a single signer \cite{Nagashima2021-ni, Osugi2012-zw}. Consequently, some studies do not evaluate model performance on unseen signers \cite{Ando2024-ow, Bai2024-yw}.
Although other studies have evaluated models on unseen signers, they relied on data of novice signers imitating JSL signs~\cite{Takazume2023-os}. This approach presents a significant limitation, as the spatial and temporal characteristics of non-native signing differ significantly from those of native signers. 


In contrast, our dataset directly addresses these limitations. 
It features 19 participants, all of whom are Deaf individuals who use JSL in their daily lives. 
Furthermore, the reviews were performed by Deaf experts, including those with expertise in sign linguistics. 
We believe this dataset provides a strong foundation for training ISLR models that can generalize to unseen native signers, a critical and previously unmet requirement for real-world applications.


\begin{figure*}[tbp]
  \begin{center}
    \includegraphics[width=0.8\linewidth]{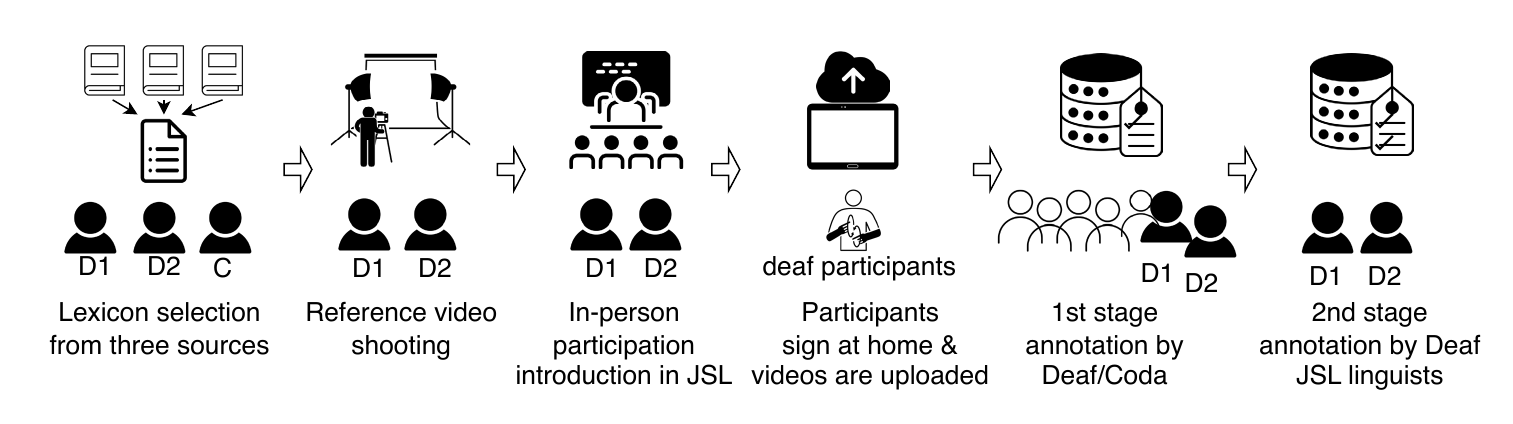}
    \vspace{-0.5cm}
    \caption{Overview of our Deaf-centric process of data collection involving deaf participants. Deaf authors D1, D2, and the Coda author C led the process.}
    \label{fig:collection_process}
    \vspace{-0.5cm}
  \end{center}
\end{figure*}
\section{Collection Methodology}
\label{sec:method}

We adopted a \textbf{Deaf-centric} approach to ensure that the dataset meaningfully serves the needs and practices of the Deaf community. 
This ``by the community, for the community'' approach is crucial for developing technology that accurately reflects its target users.
Our data collection process was \textbf{led by two Deaf authors (D1 and D2) and one Coda author (C), who are experts in Deaf education or sign language linguistics.}
D1 and D2 grew up in families with a high proportion of Deaf members. 
D1 has professional experience working with deaf children in a preschool setting and maintains daily interaction with deaf people. 
D2 possesses extensive experience teaching JSL to hearing students.
C has professional experience teaching English to deaf children at a school for the Deaf.
Moreover, participants were required to be  \textbf{deaf adults aged 18 or older who use JSL in their daily lives.}

The data collection process is illustrated in \Cref{fig:collection_process}. 
First, we selected the lexicon and then shot reference videos to instruct participants on which signs to perform.
Next, D1 or D2 explained the data collection procedure to participants in JSL.
The explanation included the data collection process using tablet devices and the associated risks of releasing the data, including the faces of individuals, to the public.
Participants took the tablet devices home to record data, which was automatically uploaded to the cloud, and returned the tablets upon completion of the data collection.
To ensure the quality and integrity of our dataset, all collected sign language videos underwent a two-stage review process conducted by Deaf or Coda individuals.
We detail each step in the following sections.



\subsection{Lexicon Selection and Shooting Reference Videos}
\label{ssec:lexicon_selection}

The lexicon selection team (D1, D2, and C) selected approximately 270 JSL words, which is 2.7 times larger than the previous largest dataset with multiple signers~\cite{Bono2014-dg}. 

\textit{Purpose of our lexicon.} Different from other ISLR datasets, our lexicon aimed to create a priority list for hearing parents striving to establish effective communication with their deaf children. Therefore, we had to create our own lexicon instead of adapting the lexicons of other sign languages.
It has been reported \cite{Takei2001-tx} that after one year of age, children begin to produce \textit{two-word sentences} by combining pointing with a sign (e.g., pointing + verb/adjective or verb/adjective + pointing), which corresponds to a pronoun + verb/adjective structure. 
Here, pointing is used as a pronoun in sign language acquisition. 
Therefore, the acquisition of verbs and adjectives, which cannot be easily substituted by pointing, should be prioritized over nouns.
As a result, our lexicon prioritizes verbs, adjectives, and greeting expressions that the researchers identified as high-frequency items for deaf children. 

\textit{Extracting Japanese words from three sources.} We referenced three sources for the lexicon.
From the 103 words listed under \textit{Action words} in the MacArthur-Bates Communicative Development Inventories \cite{Fenson2007-ac}, the lexicon selection team translated them into Japanese. For words such as ``open,'' which have distinct intransitive (\begin{CJK}{UTF8}{min}開く\end{CJK}, \textit{aku}) and transitive (\begin{CJK}{UTF8}{min}開ける\end{CJK}, \textit{akeru}) forms in Japanese, both variants were included in the translation.
Daily routines and greetings, action words, temporal expressions, and descriptive/attributive words were chosen from the Japanese MacArthur-Bates Communicative Development Inventories (CDI) \cite{Ogura2016-ql}. In addition, verbs were extracted from first-grade Japanese language textbooks \cite{Kai2015-if}. 
The list resulted in 200 Japanese words, which were selected based on their frequent usage among Japanese deaf children, as anticipated by the lexicon selection team.

\textit{Japanese to JSL translation.} 
Finally, for approximately 200 Japanese vocabulary items, D1 and D2 primarily led the process of identifying corresponding 270 JSL expressions. 
For verbs such as ``wash'' (\begin{CJK}{UTF8}{min}洗う\end{CJK}, \textit{arau}), where the sign varies depending on the direct object, 2-3 sign variants were selected. 
Consequently, multiple JSL expressions may correspond to a single Japanese word. 
Additionally, JSL-specific expressions such as (\begin{CJK}{UTF8}{min}オーバー\end{CJK}, \textit{o-ba-}) were incorporated into the list.
(\begin{CJK}{UTF8}{min}オーバー\end{CJK}, \textit{o-ba-}) carries context-dependent meanings but is frequently used in interactions with young children to express surprise, equivalent to expressions such as ``Really?'' or ``Wow, amazing!'' Additionally, the Japanese word for writing has signs corresponding to vertical and horizontal writing.





\subsection{Shooting Reference Videos}

\begin{figure}[tbp]
  \begin{center}
    \includegraphics[width=0.8\linewidth]{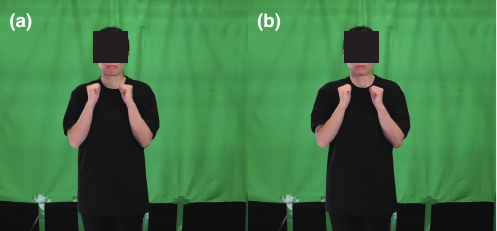}
    \caption{Sample reference videos used in the data collection. (a) is ``scary'' (\begin{CJK}{UTF8}{min}怖い\end{CJK}, \textit{kowai}), and (b) is ``cold'' (\begin{CJK}{UTF8}{min}寒い\end{CJK}, \textit{samui}), with the same body movement but different mouth movements.)} 
    \label{fig:reference_video}
  \end{center}
\end{figure}

\textit{The process.} 
We recorded D1 and D2 signing reference videos for the entire lexicon in a studio.
Another Deaf individual trimmed the video's duration. Finally, the authors discussed and selected the clearer version for all 270 vocabulary items as reference videos (\Cref{fig:reference_video}).

\textit{Apparatus.} We shot videos against a green screen using a video camera (Panasonic Lumix DC-GH7L) at 4K 60 fps.
Lighting was positioned to minimize facial shadows during signing, and signers wore plain black T-shirts to ensure that their hand shapes were clearly visible.

\subsection{In-person Participation Introduction in JSL}
\label{ssec:inperson_intro}
We held an in-person introduction in JSL, led by D1 or D2, to ensure that participants understood the data collection process using tablets and the associated risks of releasing the data, including the faces of individuals, to the public.

After obtaining consent in JSL, each participant was asked to perform 270 words 20 times each, totaling 5400 examples, at their favorite place and time.
They were informed that they would receive 75,000 yen in compensation for performing 5,400 signs, with compensation provided proportionally based on the number of completed signs if they quit earlier.
To prevent leakage of residential location and other personal information during data publication, we instructed participants to ensure that nothing revealing such information appeared in the background and that only the participant themselves was visible in the recordings.
Additionally, we requested that participants keep their hands within the camera's field of view throughout the recording.
Participants contacted D1, D2, or C if they had questions or problems during the collection period.

\subsection{JSL Recorder App}




Recording was performed using an open-source sign language recording application installed into a Google Pixel Tablet with a built-in RGB camera. 

The video resolution is 2560 x 1920 pixels, and the frame rate is 30 fps with an average bitrate of 15 Mbps.

\begin{figure}[tbp]
  \begin{center}
    \includegraphics[width=0.8\linewidth]{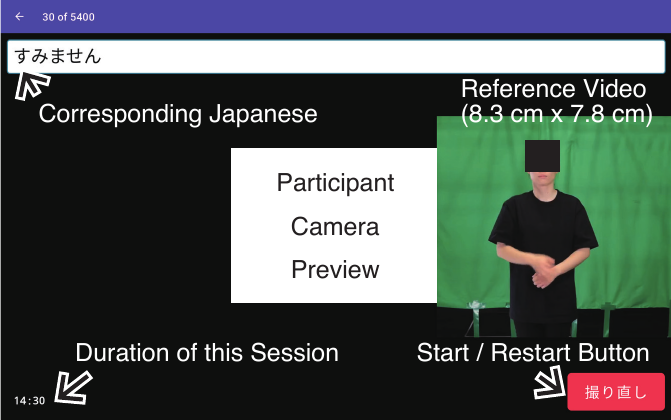}
    \caption{Screenshot of the JSL recorder tablet application. The participant swipes to the left after finish signing each example.)}
    \label{fig:rth}
    \vspace{-0.5cm}
  \end{center}
\end{figure}

From the app's home screen, pressing the session start button transitions to the sign recording screen, which displays the corresponding Japanese word, the participant's video in the center, and the reference video on the right, as shown in \Cref{fig:rth}.
D1 and D2 verified that the reference video is sufficiently large (8.3 × 7.8 cm on the tablet) for deaf participants to understand which sign to perform.
Participants could check the reference video, press the start button on screen, perform the sign, and swipe left when finished to proceed to the next sign.
If participants wanted to redo a sign, they could press the restart button to retry.
To maintain high signing quality, we limited each session to 30 examples and ensured that the same sign did not appear more than once within a single session.

Since tablet devices cannot instantly start recording, the recording began at the start of the session. 
We separately recorded timestamps for when the start button was pressed and when participants swiped after completing each sign, marking the start and end times of each sign.
Based on these records, we segmented each sign in the video.

We found that the 5400th example (the sign corresponding to ``call someone'') could not be recorded due to an application bug.
This issue was experienced by all participants who finished all recordings.

\subsection{Review Procedure}


To ensure the quality and integrity of our dataset, all collected sign language videos underwent a rigorous, two-stage review process conducted \textbf{by native signers and linguistic experts}. This pipeline was designed to filter out erroneous data, protect privacy, and categorize valid sign variations.

\textit{Two-Stage review pipeline.}
The review process was divided into two check steps. 
The entire review team performed an initial quality assessment to filter the collected signs.
Signs that passed the initial check (or were flagged as ``Unsure'' ) were escalated to a 2nd check. 
This stage was conducted by Deaf linguists inside the review team.

As shown in \Cref{fig:rth}, our data collection app presents participants with a corresponding Japanese gloss and a reference video simultaneously. 
In this way, participants might perform a different sign from the one shown in the reference video, just by looking at the Japanese gloss, which is a common challenge in collecting sign language datasets~\cite{Hochgesang2018-ea}.
Our review process with Deaf linguists allowed us to filter the dataset, retaining only the signing words that have the same meaning as those shown in the reference video.

\section{Dataset Composition}
\label{sec:dataset}

\subsection{Statistics}
Twenty \textbf{Deaf individuals who use JSL daily} participated in data collection by snowball sampling. 
Still, one was excluded from the dataset due to frequent unnatural signing identified during the review process. 
Therefore, the dataset contains data from a total of 19 individuals, consisting of 11 males and 8 females, ranging in age from 18 to the 70s.
The dataset was collected in Japan with an entirely Japanese participant pool. 

Participants recorded videos in various locations and at different times of day, resulting in diverse backgrounds and lighting conditions.
Our dataset comprises 36,705 videos that passed our two-stage review pipeline.
Surprisingly, 9.8\% of clips in the confusable glosses were removed as variants: the meaning differs from the reference signing, suggesting that a careful review is necessary to keep the samples for each gloss clean. 
The sign corresponding to the word ``get angry'' had the most data (309 samples), while the sign corresponding to the word ``think'' had the fewest (27 samples). 
We will also publicly release the reference videos used during data collection, an example of which is shown in \Cref{fig:reference_video}.


We split the dataset into train, validation, and test sets. The signers were divided into 15 (train), 2 (val), and 2 (test), with gender balance maintained across splits. This technique ensures a disjoint signer setup (i.e., individuals in one split do not appear in any other). 
All 270 sign classes are represented in each split. The resulting data proportions are 81.3\% (train), 8.4\% (validation), and 10.3\% (test).

%

\subsection{Descriptions for Distinguishing Confusable Signs}
We provide additional descriptions to distinguish confusable signs and guide researchers in creating classifiers for them.
D2 added descriptions for pairs that cannot be distinguished by handshape alone, together with an explanation of how to distinguish each pair. C verified the descriptions.
Twenty-two glosses belong to this set, differing in aspects such as mouthing or eye movement.
Two groups rely partly on context, however, most pairs in our glosses can be distinguished by a Deaf person without the surrounding context.

\begin{figure}[tbp]
  \begin{center}
    \includegraphics[width=0.8\linewidth]{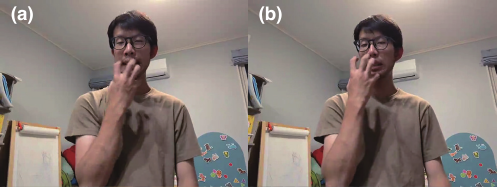}
    \caption{Frequently confused examples. (a) is ``bitter (taste)'' (\begin{CJK}{UTF8}{min}苦い\end{CJK}, \textit{nigai}), and was misclassified as ``spicy'' (\begin{CJK}{UTF8}{min}辛い\end{CJK}, \textit{karai}), which is (b). They could be classified by the mouth movements.}
    \label{fig:confused}
    \vspace{-0.5cm}
  \end{center}
\end{figure}
We illustrate challenging and easily confused examples from our dataset in \Cref{fig:confused}. 
For instance, (a) shows the sign ``bitter'' (taste), and (b) is an example of ``spicy''. 
They could be distinguished by their mouth movements. 

Using these descriptions, sign language recognition researchers can refine the encoder they use or guide the model's attention.
We consider such labeling a useful guide for improving accuracy even with limited data, and we present an example of its use in \cref{ssec:showcase}, demonstrating that it improves recognition accuracy.

\section{Experiments}
\label{sec:experiments}

\begin{table*}[ht]
\centering
\footnotesize
\begin{tabular}{lcccccccc}
\toprule
& & & \multicolumn{2}{c}{All (270 glosses)} & \multicolumn{2}{c}{Confusable (22 glosses)} & \multicolumn{2}{c}{Others (248 glosses)} \\
\cmidrule(lr){4-5} \cmidrule(lr){6-7} \cmidrule(lr){8-9}
Method & input & MACs (G) & Top-1 & F1 & Top-1 & F1 & Top-1 & F1 \\
\midrule
SPOTER~\cite{Bohacek2022-bp}            & keypoint & 0.38 & 78.7 & 78.5 & 51.7 & 50.9 & 81.1 & 80.9 \\
PopSign ASL v2.0~\cite{Neubauer2026-ru} & keypoint & 1.6 & 81.9 & 81.4 & 55.3 & 50.1 & 84.3 & 84.2 \\
MASA~\cite{Zhao2024-kv}                 & keypoint & 3.9 & 82.8 & 83.0 & 54.8 & 51.1 & 85.3 & 85.8 \\
NLA-SLR~\cite{Zuo2023-qs}               & keypoint & 55.7 & 91.6 & 91.9 & 61.2 & 58.1 & \textbf{94.3} & \textbf{94.9} \\
NLA-SLR + mouthing (ours)               & keypoint & 58.0 & \textbf{92.2} & \textbf{92.5} & \textbf{68.3} & \textbf{65.4} & \textbf{94.3} & \textbf{94.9} \\
\midrule
I3D~\cite{Carreira2017-ne}              & RGB & 111.2 & 83.5 & 83.1 & 54.1 & 47.7 & 86.1 & 86.2 \\
NLA-SLR~\cite{Zuo2023-qs}               & RGB & 71.9 & 91.9 & 92.1 & 57.6 & 53.5 & \textbf{94.9} & \textbf{95.5} \\
NLA-SLR + mouthing (ours)               & RGB & 74.7 & \textbf{92.5} & \textbf{92.9} & \textbf{65.7} & \textbf{63.3} & \textbf{94.9} & \textbf{95.5} \\
\bottomrule
\end{tabular}
\caption{Results across confusable vs.\ other glosses.}
\label{tab:results}
\end{table*}

To clarify the characteristics and challenges of our dataset, we evaluated the recognition accuracy using existing ISLR models. We selected I3D~\cite{Carreira2017-ne} and NLA-SLR (RGB input)~\cite{Zuo2023-qs} as a pixel-based method and SPOTER~\cite{Bohacek2022-bp}, PopSign ASL v2.0~\cite{Neubauer2026-ru}, MASA~\cite{Zhao2024-kv}, and NLA-SLR (keypoint input)~\cite{Zuo2023-qs} as pose-based methods.
\Cref{tab:results} reports the Top-1 accuracy and F1 score over the 270 glosses, together with the scores on the 22 confusable glosses determined by the linguists and the remaining 248 glosses.
Although almost every pair in our dataset could be distinguished by a Deaf person without the surrounding context, we found that even state-of-the-art ISLR models, which achieve an average accuracy above 90\%, still struggle to distinguish the glosses in the confusable set.


\begin{figure}[tbp]
  \begin{center}
    \includegraphics[width=\linewidth]{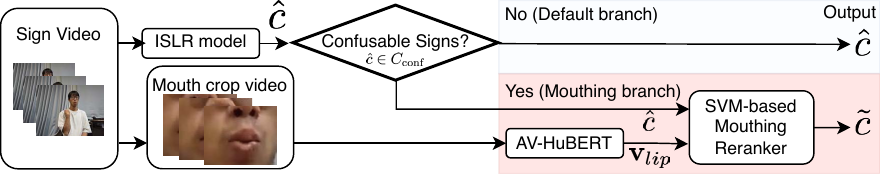}
    \caption{The proposed mouthing-based reranker to classify the confusable sets.}
    \label{fig:reranker}
    \vspace{-0.5cm}
  \end{center}
\end{figure}
\label{ssec:showcase}
As an example that leverages the Deaf linguist's description of pairs that cannot be distinguished by handshape alone, we also propose the model shown in \cref{fig:reranker}, which adds a mouthing classifier to NLA-SLR.
To extract mouth-movement features, we used AV-HuBERT (Audio-Visual Hidden Unit BERT)~\cite{Shi2022-sk}, which can perform the lip-reading task.
Among the 22 confusable glosses, 16 glosses were identified as distinguishable by mouth movement according to the linguist-derived description. 
During training, we built one SVM per confusable pair among the 16 glosses to discriminate within each confusable combination based on mouth movement.
At test time, when the predicted sign from the ISLR model is one of the 16 glosses, we extract mouth-region features with AV-HuBERT, classify with the SVM for that pair, and produce the final output.

We found that this raises the overall accuracy over NLA-SLR, which is typically the most accurate among existing models, while improving the F1 and Top-1 scores, especially for confusable glosses.
\cref{tab:rerank_diff} shows the change in Top-1 accuracy and F1 score when mouthing is added to NLA-SLR (RGB).
These results show that some confusable signs improve in score, while others do not improve.
For the other 254 glosses, there was no change in accuracy or F1 score.
A Wilcoxon signed-rank test over the 270 glosses showed that our model achieved a significantly higher F1 score than NLA-SLR ($p=0.009$ for the keypoint-based model and $p=0.008$ for the RGB-based model).



Therefore, we found that mouth-movement features from AV-HuBERT alone can increase accuracy on the confusable glosses.
These results suggest that linguist-derived descriptions can be used to build better models, and incorporating mechanisms that attends to other cues such as eye movements and hand speed may further improve performance.

\begin{table}[ht]
\centering
\scriptsize
\begin{tabular}{l cc}
\toprule
Gloss & $\Delta$Top-1 & $\Delta$F1 \\
\midrule
Thank you for the meal. {\tiny (s0006\_itatakimasu)} & $-19.0$ & $-3.5$ \\
Thank you for the meal. {\tiny (s0033\_kochisousamateshita)} & $\mathbf{+31.2}$ & $\mathbf{+38.5}$ \\
\midrule
scared {\tiny (s0034\_kowai\_Bu\_i\_)} & $\mathbf{+37.5}$ & $\mathbf{+33.8}$ \\
cold {\tiny (s0036\_samui\_Han\_i\_)} & $\mathbf{+10.0}$ & $\mathbf{+15.5}$ \\
cold {\tiny (s0057\_tsumetai\_Leng\_tai\_)} & $\mathbf{+44.4}$ & $\mathbf{+43.9}$ \\
\midrule
spicy {\tiny (s0024\_karai\_Xin\_i\_)} & $0.0$ & $+0.8$ \\
bitter {\tiny (s0064\_nikai\_Ku\_i\_)} & $\mathbf{+11.1}$ & $\mathbf{+20.0}$ \\
laugh {\tiny (s0091\_warau\_Xiao\_u\_)} & $0.0$ & $0.0$ \\
\midrule
arrive {\tiny (s0056\_tsuku\_Zhao\_ku\_)} & $\mathbf{+12.5}$ & $\mathbf{+24.6}$ \\
stop {\tiny (s0058\_tomaru\_Zhi\_maru\_)} & $\mathbf{+50.0}$ & $\mathbf{+41.8}$ \\
\midrule
pour {\tiny (s0047\_sosoku)} & $0.0$ & $0.0$ \\
pour {\tiny (s0050\_sosoku\_Zhu\_ku\_)} & $0.0$ & $0.0$ \\
\midrule
gather {\tiny (s0039\_shiyuukousuru\_Ji\_He\_suru\_)} & $0.0$ & $0.0$ \\
small {\tiny (s0052\_chiisai\_Xiao\_sai\_)} & $0.0$ & $0.0$ \\
\midrule
happy {\tiny (s0010\_ureshii\_Xi\_shii\_)} & $0.0$ & $0.0$ \\
fun {\tiny (s0052\_tanoshii\_Le\_shii\_)} & $0.0$ & $0.0$ \\
\bottomrule
\end{tabular}
\caption{Change in Top-1 accuracy and F1 (percentage points) after the mouthing rerank, for confusable pairs distinguished mainly by mouthing/mouth cues.}
\label{tab:rerank_diff}
\end{table}




\section{PopSign JSL}
\begin{figure}[tbp]
  \begin{center}
    \includegraphics[width=0.6\linewidth]{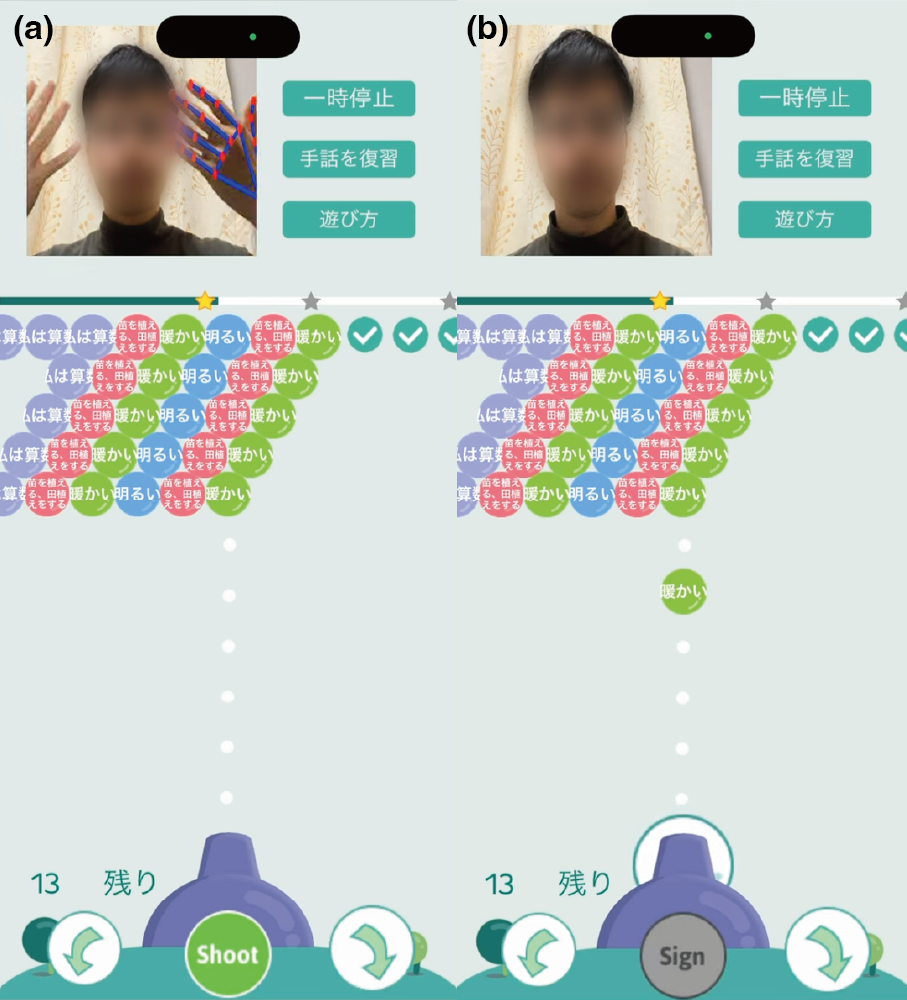}
    \caption{A screenshot of the playable version of PopSign JSL, where (a) the user is signing ``warm'' (\begin{CJK}{UTF8}{min}暖かい\end{CJK}, \textit{atatakai}) and hand landmarks are recognized by MediaPipe (only the left hand is visualized). The landmarks are fed into a Bi-LSTM model trained with our dataset. (b) The bubble labeled with the recognition result is released.}
    \label{fig:popsign}
    \vspace{-0.5cm}
  \end{center}
\end{figure}

We are currently working with the Deaf community to determine which types of games might be most compelling for learning JSL vocabulary. However, a local Deaf school  is interested in a port of the open source PopSign ASL game \cite{Starner2023-vt} to help their Japanese hearing parents of Deaf children learn how to better communicate with their children in JSL. In addition, the Deaf school has requested a version of PopSign to help their ninth-grade Deaf class learn written English vocabulary. The intention is that the children will sign the JSL translation of a concept when cued with a written English word.  

PopSign JSL is similar to Taito's popular 1994 Japanese arcade game Puzzle Bobble, known internationally as Bust-A-Move. In the original game, players are presented with a field of colored balls. Players are provided with a colored ball at each turn, which they must release into the field of balls. Whenever three or more balls of the same color touch, they burst and disappear from the screen. The player attempts to remove all balls from the screen.  PopSign JSL modifies this game mechanic in that the player can select the color of the ball they wish to release by signing the JSL sign associated with that color. Each color ball is labeled with a written concept that the player must match with their signing. Poor signing results in the wrong color ball being released. 

Figure~\ref{fig:popsign} shows a playable version of PopSign JSL designed for a smartphone. While many recognition systems tested here are not suitable for smartphones, the 3-layer Bi-LSTM recognizer runs in real time and achieves sufficient accuracy for gameplay. For hearing parents learning JSL, the bubbles are labeled with Japanese text. For the Japanese Deaf students learning English text, the balls are labeled in English. Future work includes user testing and iteration of the game, as well as increasing the vocabulary size. 

\section{Ethics and Safety Discussion}
\label{sec:ethics}
This research was conducted with Deaf researchers at the core of the team, maintaining constant awareness of the Deaf community.
This research, including the dataset publication, was conducted after receiving approval from the institutional review board of the authors' organizations.
Participants received 75,000 yen in compensation for performing 5,400 signs, which took an average of 8.7 hours. 
Four participants did not complete the data collection, but compensation was provided based on the number of completed signs, and their videos were included in the dataset.

The dataset will be released under a Creative Commons CC-BY 4.0 license, allowing both academics and industry to use it.
A key agreement of this license is that any use must not infringe upon the participants' moral rights, particularly any application that is prejudicial to their honor or reputation.
Participants retain the right to withdraw their data at any time. They can have their videos removed from our servers at any time by signing and submitting the withdrawal of consent form provided to them. However, we have explicitly informed participants that we cannot enforce the deletion of data that may have been downloaded by third parties after the dataset's public release at camera-ready.
Our two-stage review process rejected videos that contained privacy issues, such as visible personal information.

While the signs in the reference videos for this collected data were selected through discussion to represent the most commonly used forms among various sign variants, it should be noted that this dataset should not be used for establishing a \textit{standard} for JSL, though there is a risk users might use it in that way.


\section{Limitations and Future Work}
\label{sec:limitations}

\textit{Coverage of JSL variations.}
JSL varies by region and community, such as among different schools and communities.
However, this dataset does not aim to cover all such variations, as our first application is not JSL-to-Japanese translation. Rather, our first future application is a sign language learning app, designed to teach one variant of a sign to facilitate communication between hearing parents and deaf children. 
Another application of ISLR would be helping JSL signers look up Japanese words in JSL to help them learn Japanese, which is a second language for them.
In these cases, it would be desirable to support various JSL styles, including dialectal signs.

\textit{Word-level lexicon.}
This research focuses on individual words, especially verbs and adjectives, as they are essential for communication between hearing parents and deaf children in their early stages, as discussed in \Cref{ssec:lexicon_selection}.
Future collection with more nouns, phrases, or sentences will be necessary to create JSL-Japanese translation models.

\textit{Data biases.}
All participants in this dataset collection were Japanese. 
As a result, the dataset primarily represents a single ethnic group and does not capture a wide range of skin tones.
Consequently, pixel-based ISLR methods trained on this data may struggle to generalize to different skin tones. 
However, pose-based methods are expected to be robust to such variations. Furthermore, participants were recruited via snowball sampling, which means we cannot rule out potential signing biases (e.g., community-specific variations) introduced by this sampling method. 
Finally, individuals under 18 are not included in this dataset due to the difficulties associated with obtaining consent for the public release of data that includes their faces.


\section{Conclusion}
\label{sec:conclusion}


In this paper, we introduced JSL-DC, the largest JSL dataset by video count for ISLR to date, with linguist-derived descriptions to distinguish confusable signs. 
Our experiments showed that our model, inspired by the description, improved the F1 score from 53.5\% to 63.3\% compared to the state-of-the-art method on our confusable glosses.
We believe this data will enable various applications, starting with sign language learning for families with deaf children to support their language development, and ultimately bring a positive impact to the Deaf community in Japan.
{
    \small
    \bibliographystyle{ieeenat_fullname}
    \bibliography{paperpile}
}


\end{document}